\documentclass[runningheads]{llncs}

\usepackage{cite,amsmath,amssymb,amsfonts,algorithmic,graphicx,textcomp,comment}
\usepackage{times}
\usepackage{epsfig}
\usepackage[ruled,vlined]{algorithm2e}
\usepackage{enumerate}
\usepackage{amsmath}
\usepackage{amssymb}
\usepackage{fancyhdr}
\usepackage{comment}
\usepackage{subfigure,pifont}
\usepackage{todonotes}
\usepackage{makecell}
\usepackage{floatrow}
\usepackage{subfig}
\usepackage{delarray} 
\usepackage[inline]{enumitem}
\usepackage[capitalize]{cleveref}
\usepackage{tabularx, booktabs, makecell,multirow, multicol,pifont, multirow,caption, threeparttable}

\newcommand{\cmark}{\ding{51}}%
\newcommand{\xmark}{\ding{55}}%

\newfloatcommand{capbtabbox}{table}[][\FBwidth]

\newcommand{\myfirstpara}[1]{\noindent \textbf{#1}:}
\newcommand{\mypara}[1]{\vspace{0.5em} \myfirstpara{#1}}

\usepackage{xcolor, soul}
\definecolor{todocolor}{RGB}{200,120,120}
\sethlcolor{todocolor}

\begin{document}
\title{Efficient and Generic Interactive Segmentation Framework to Correct Mispredictions \\ during Clinical Evaluation of Medical Images}

\titlerunning{A Generic Interactive Segmentation Framework}
\author{Bhavani Sambaturu\inst{1} % index{Sambaturu, Bhavani} \and
Ashutosh Gupta\inst{2} % index{Gupta, Ashutosh} \and
C.V. Jawahar\inst{1} % index{C.V., Jawahar} \and
Chetan Arora \inst{2}} % index{Arora, Chetan}}
\authorrunning{B. Sambaturu et al.}
\institute{International Institute of Information Technology, Hyderabad, India 
\email{bhavani.sambaturu@research.iiit.ac.in}\\
\and
Indian Institute of Technology, Delhi, India
}

\maketitle              

\begin{abstract}
Semantic segmentation of medical images is an essential first step in computer-aided diagnosis systems for many applications.  However, given many disparate imaging modalities and inherent variations in the patient data, it is difficult to consistently achieve high accuracy using modern deep neural networks (DNNs).
This has led researchers to propose interactive image segmentation techniques where a medical expert can interactively correct the output of a DNN to the desired accuracy. However, these techniques often need separate training data with the associated human interactions, and do not generalize to various diseases, and types of medical images. In this paper, we suggest a novel conditional inference technique for DNNs which takes the intervention by a medical expert as test time constraints and performs inference conditioned upon these constraints. Our technique is generic can be used for medical images from any modality. Unlike other methods, our approach can correct multiple structures simultaneously and add structures missed at initial segmentation. We report an improvement of 13.3, 12.5, 17.8, 10.2, and 12.4 times in user annotation time than full human annotation for the nucleus, multiple cells, liver and tumor, organ, and brain segmentation respectively. We report a time saving of 2.8, 3.0, 1.9, 4.4, and 8.6 fold compared to other interactive segmentation techniques.  Our method can be useful to clinicians for diagnosis and post-surgical follow-up with minimal intervention from the medical expert. The source-code and the detailed results are available here \cite{project-page}.


\end{abstract}

\section{Introduction}
\label{sec:introduction}

\begin{table*}[ht]
%\small
	\centering
	\setlength{\tabcolsep}{1.5pt}
	\begin{tabular}{lm{6.5cm}cccccc}
		\toprule[1.5pt]
		Capability  & Description & \cite{brs} & \cite{fbrs} & \cite{dextr} & \cite{nuclick} & Ours \\
\midrule[1pt]      
		\multirow{3}{*}{\makecell[l]{Feedback \\ mode}} 
		& Point  & \cmark & \cmark & \xmark & \cmark & \cmark \\
		& Box & \xmark & \xmark & \cmark & \xmark & \cmark \\
		& Scribble & \xmark & \xmark &  \xmark & \xmark & \cmark \\
\midrule
		\multirow{2}{*}{\makecell[l]{Training \\ Requirement}} &
		Pre-training with user interaction & \cmark &\cmark & \cmark & \cmark & \xmark\\
		& Can work with any pre-trained DNN & \cmark & \cmark & \xmark & \xmark & \cmark \\
\midrule
		\multirow{2}{*}{\makecell[l]{Correction \\ Modes}} &
		Correct multiple labels & \xmark & \xmark & \xmark & \xmark & \cmark \\
		& Insert missing labels & \xmark & \xmark & \xmark & \xmark & \cmark \\
\midrule
		\multirow{2}{*}{Generalization} &
		Adapt: Distribution mismatch & \xmark & \xmark & \xmark & \xmark & \cmark \\
		& Segment new organs than trained for & \xmark & \xmark & \xmark & \xmark & \cmark \\
\bottomrule[1.5pt]
	\end{tabular}
	\caption{Comparative strengths of various interactive segmentation techniques.}
	\label{tab:sota-comparison}
%	\vspace{-1em}
\end{table*}

%\begin{figure}[t]
%	\setlength\abovecaptionskip{-5pt}
%	\includegraphics[width=\linewidth, height=0.62\linewidth]{images/teaser2}
%	\caption{The figure shows the use of our algorithm in different modalities. We first obtain the initial segmentation with a pre-trained model. The user then provides scribbles where he desires improvement. The network then uses the scribble information to provide an improved segmentation. To demonstrate the generality of our technique we show the results on 3 different input modalities: Nuclei segmentation for the Monuseg challenge \cite{monuseg} for 2D Microscopy (top row), liver and tumor segmentation from CT images of the LiTS challenge \cite{challenlitsge} (middle row), and the CHAOS MRI segmentation from MRI images (bottom row).}
%	\label{fig:teaser}
%\end{figure}

\myfirstpara{Motivation}
Image segmentation is a vital imaging processing technique to extract the region of interest (ROI) for medical diagnosis, modeling, and intervention tasks. It is especially important for tasks such as the volumetric estimation of structures such as tumors which is important both for diagnosis and post-surgical follow-up.
%The process involves segmenting various body organs and tissues for boundary detection, tumor detection and segmentation, as well as mass detection which is then used in applications such automatic measurement of organs, cell counting, or simulations based on the extracted boundary information. The result of the segmentation is typically used to obtain further diagnostic insights in different downstream applications. 
%
A major challenge in medical image segmentation is the high variability in capturing protocols and modalities like X-ray, CT, MRI, microscopy, PET, SPECT, Endoscopy and OCT. Even within a single modality, the human anatomy itself has significant variation modes leading to vast observed differences in the corresponding images. Hence, fully automated state-of-the-art methods have not been able to consistently demonstrated desired robustness and accuracy for segmentation in clinical use. This has led researchers to develop techniques for interactive segmentation which can correct the mispredictions during clinical evaluation and make-up for the shortfall.

\mypara{Current Solutions}
Though it is helpful to leverage user interactions to improve the quality of segmentation at test time, this often increases the burden on the user. A good interactive segmentation method should improve the segmentation of the image with the minimum number of user interactions. Various popular interactive segmentation techniques for medical imaging have been proposed in the literature \cite{rother2004grabcut,geos,slicseg}. The primary limitation is that it can segment only one structure at a time. This leads to a significant increase in user interactions when a large number of segments are involved. Recent DNN based techniques \cite{brs,dextr,fbrs} improve this aspect by reducing user interactions. It exploits pre-learnt patterns and correlations for correcting the other unannotated errors as well. However, they require vast user interaction data for training the DNN model, which increases cost and restricts generalization to other problems. 

%Hence, most interactive segmentation networks leverage a semantic segmentation architecture as a backbone. Additionally, these techniques typically incorporate user interactions in the following manner:
%\begin{enumerate}
%\item Take the foreground and background clicks, create a euclidean or geodesic distance map and put it as an additional channel to the input image and train an existing semantic segmentation network such as DeepLab in an early fusion manner 
%\item Put an additional refinement network to the exiting semantic segmentation network. 
%\end{enumerate}
%Though deep learning based approaches have shown a better performance in comparison to the other styles \cite{3d-grabcut,growcut,geos,slicseg} methods, they still suffer from similar limitations such their ability to correct only one structure at a time. Thus, the reduced efficiency, and the annotated data requirement restrict the practical application of many of these techniques. 

\mypara{Our Contribution}
We introduce an interactive segmentation technique using a pre-trained semantic segmentation network, without any additional architectural modifications to accurately segment 2D and 3D medical images with help from a medical expert. Our formulation models user interactions as the additional test time constraints to be met by the predictions of a DNN. The Lagrangian formulation of the optimization problem is solved by the proposed alternate maximization and minimization strategy, implemented through the stochastic gradient descent. This is very similar to the standard back-propagation based training for the DNNs and can readily be implemented. The proposed technique has several advantages:
\begin{enumerate*}[label=(\arabic*)]
	\item exhibits the capability to correct multiple structures at the same time leading to a significant reduction in the user time. 
	\item exploits the learnt correlations in a pre-trained deep learning semantic segmentation network so that a little feedback from the expert can correct large mispredictions.
	\item requires no joint training with the user inputs to obtain a better segmentation, which is a severe limitation in other methods \cite{brs,fbrs}. 
	\item add missing labels while segmenting a structure if it was missed in the first iteration or wrongly labeled as some other structure. The multiple types of corrections allow us to correct major mispredictions in relatively fewer iterations.
	\item handle distribution mismatches between the training and test sets. This can arise even for the same disease and image modality due to the different machine and image capturing protocols and demographies. 
	\item for the same image modality, using this technique one can even segment new organs using a DNN trained on some other organ type. 
\end{enumerate*}
\cref{tab:sota-comparison} summarizes the comparative advantages of our approach.

\section{Related Work} \label{related_work}
%\vspace{-1.1em}
\myfirstpara{Conventional Techniques}
 Interactive segmentation is a well-explored area in computer vision and some notable techniques are based on Graph Cuts \cite{rother2004grabcut, straehle2011carving, geos}, Edge or Active Contours \cite{Kass1988, top2010spotlight}, Label propagation using Random Walk or other similar style \cite{Grady2006, slicseg}, and region-based methods \cite{Sahoo88, Horowitz1976}. In these techniques, it is not possible to correct multiple labels together without the user providing the initial seeds and also not possible to insert a missing label. 

\mypara{DNN based Techniques}
DNN based techniques use inputs such as clicks \cite{brs,fbrs}, scribbles \cite{scribblesup}, and bounding boxes \cite{dextr} provided by a user. Other notable techniques include \cite{ZLiCVPR2018, XuCVPR2016,  AcunaCVPR2018, dextr, fbrs}. These methods require special pre-training with user-interactions and associated images. This increases the cost of deployment and ties a solution to pre-decided specific problem and architecture. 

\mypara{Interactive Segmentation for Medical Images}
Interactive Segmentation based methods, especially for medical image data, have been proposed in \cite{deepigeos, bifseg, nuclick, uinet, scribble2label}. The methods either need the user inputs to be provided as an additional channel with the image \cite{uinet} or need an additional network to process the user input \cite{deepigeos}. BIFSeg \cite{bifseg} uses the user inputs at test time with a DNN for interactive segmentation of medical images. However, our method is significantly different in the following manner: (a) DNN - use their own custom neural networks \cite{bifseg}. However, our method can use pre-existing segmentation networks. This allows our method to use newer architectures which may be proposed in the future as well. (b) Optimization - use CRF-based regularization for label correction \cite{bifseg}. We propose a novel restricted Lagrangian-based formulation. This enables us to do a sample specific fine-tuning of the network, and allows our method to do multiple label corrections in a single iteration which is novel. (c) User Inputs - use scribbles and bounding boxes as user inputs \cite{bifseg}. We can correct labels irrespective of the type of user input provided. 

%DeepIGeos \cite{deepigeos} uses two convolutional neural network (CNN) based architecture, where first CNN takes an image and gives the initial segmentation, whereas second CNN takes a segmented image with user interactions and gives final segmentation. 

%Scribble2Label \cite{scribble2label} utilizes the user-provided scribbles along with pseudo-labels in a weakly supervised manner to segment the nuclei against the background using the U-Net architecture.

%NuClick \cite{nuclick} proposes a method for obtaining guiding signals obtained from clicks inside the nuclei from gland masks and uses multiple blocks for extracting nuclei from a range of scales. The method segments the nucleus when the user clicks inside a nucleus. 
%UI-Net \cite{uinet} uses a Fully Convolutional Network and incorporates user inputs as an additional channel. The authors have proposed a rule-based model during training to randomly select the set of seed pixels.

\section{Proposed Framework} \label{approach}

The goal is to design an approximate optimization algorithm that can encode the constraints arising from user-provided inputs in the form of scribbles. A simple gradient descent strategy similar in spirit to the Lagrangian relaxation proposed by \cite{Lee2019} is optimized. The strategy allows us to use existing libraries and infrastructure built for any image modality optimizing the loss for the DNNs using the standard back-propagation procedure. 

\mypara{Problem Definition} 
A neural network with $N$ layers is parameterized by weights $W$ from input to output. We represent this as a function $\Psi (x,y,W) \rightarrow \mathbb{R}_{+}$ to measure the likelihood of a predicted output $y$ given an input $x$ and parameters/weights $W$. We also want to enforce that the output values belong to a set of scribbles $\mathbb{S}^x$ provided by the user to correct the segmentation dependent on $x$. Here, $\mathbb{S}^x$ encodes both the location in the image where correction is required and the desired segmentation class label.  

We can express the constraint, $y \in \mathbb{S}^x$, as an equality constraint, using a function $g(y,\mathbb{S}^x) \rightarrow \mathbb{R}_+$. This function  measures the compatibility between the output $y$ and scribbles $\mathbb{S}^x$ such that $g(y,\mathbb{S}^x) = 0$ if and only if there are no errors in $y$ with respect to $\mathbb{S}^x$. In our case, $g(y,\mathbb{S}^x)$ is the cross-entropy loss between the predicted labels $y$ and the segmentation class label encoded in $\mathbb{S}^x$. This allows us to solve the optimization problem by minimizing the following Lagrangian:
\begin{equation}
\underset{\lambda}{\min} ~ \underset{y}{\max} ~ \Psi(x, y, W) + \lambda ~ g(y, \mathbb{S}^x).
\label{eq:1}
\end{equation}

\begin{figure*}[t]
	%trim={<left> <lower> <right> <upper>}
	\includegraphics[trim={0 5cm 0 3cm},clip,width=\linewidth]{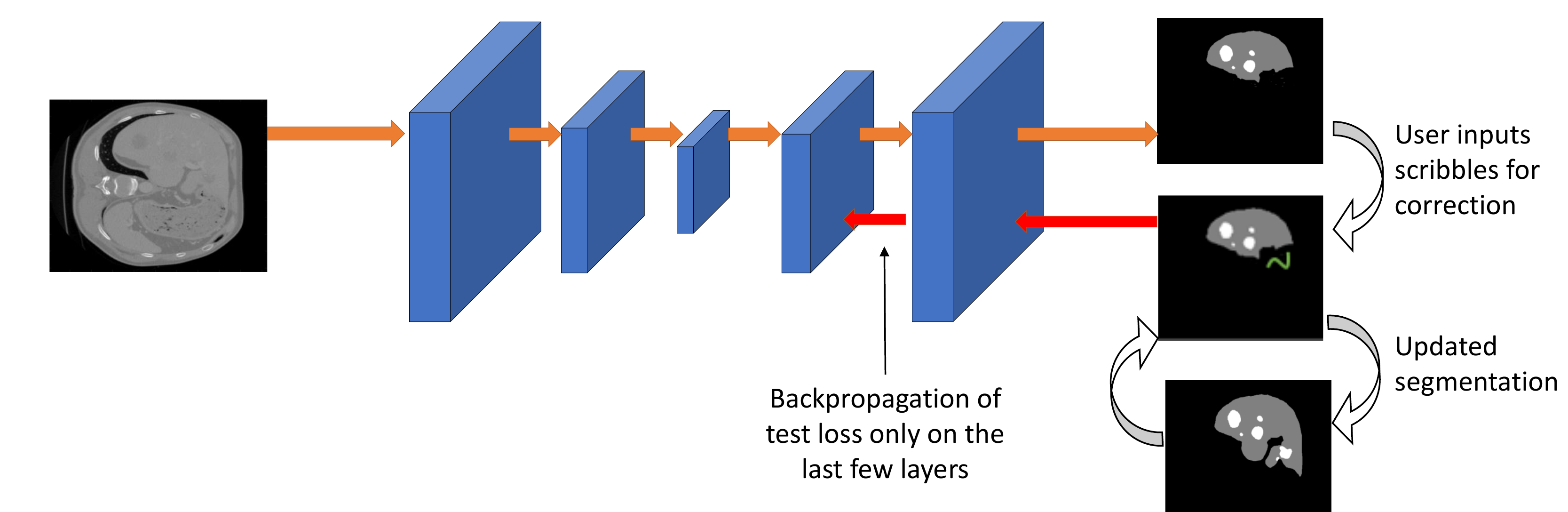}
	\caption{The figure shows the working of our algorithm. Note that depending upon the application, our framework can use different pre-trained network architectures. Hence we do not give the detailed architecture of any particular model. The first step in our framework is to obtain an initial segmentation using the pre-trained deep learning network. The user then examines the segmentation and adds scribbles where the desired correction is required. This is then used to refine the weights of the network and the improved segmentation is obtained.}
	\label{fig:proposed_framework}
\end{figure*}

Note that the compatibility constraints in $g(y, \mathbb{S}^x)$ factorize over the pixels and one trivial solution of the optimization problem as described above is to simply change the output variables to the class labels provided by the scribbles. However, this does not allow us to exploit the neighborhood information inherent in the images, and the correlations learnt by a DNN due to prior training over a large dataset. 

We note that the network's weights can also control the compatibility of the output configurations with the scribble input. Since the weights are typically tied across space, the weights are likely to generalize across related outputs in the neighborhood. This fixes the incompatibilities not even pointed-to by the limited scribbles given by the user. Hence, we propose to utilize the constraint violation as a part of the objective function to adjust the model parameters to search for an output satisfying the constraints efficiently. 

We propose to optimize a ``dual'' set of model parameters $W_{\lambda}$ over the constraint function while regularizing $W_{\lambda}$ to stay close to the original weights $W$. The network is divided into a final set of layers $l$ and an initial set of layers $N-l$. We propose to optimize only the weights corresponding to the final set of layers $W_{\lambda_{l}}$. The optimization function is given as:
\begin{equation}
\underset{{W_{\lambda_{l}}}}{\min} ~ \Psi(x, \hat{y}, W_{\lambda_{l}}) ~ g(\hat{y}, \mathbb{S}^x) + \alpha \lvert \lvert W_{l} - W_{\lambda_{l}} \rvert \rvert,
\label{eq:2}
\end{equation}
where $\hat{y} = \underset{y}{\text{arg max}} ~ \Psi(x,y,W_{\lambda_{l}})$. This function is reasonable by definition of the constraint loss $g(\cdot)$, though it deviates from the original optimization problem, and the global minima should correspond to the outputs satisfying the constraints. If we initialize  $W_{\lambda} = W$, we also expect to find the high-probability optima. If there is a constraint violation in $\hat{y}$, then $g(\cdot) > 0$, and the following gradient descent procedure makes such $\hat{y}$ less likely, else $g(\cdot) = 0$ and the gradient of the energy is zero leaving $\hat{y}$ unchanged.

The proposed algorithm (see \cref{algo}) alternates between maximization to find $\hat{y}$ and minimization w.r.t. $W_{\lambda_{l}}$ to optimize the objective. The maximization step can be achieved by employing the neural network's inference procedure to find the $\hat{y}$, whereas minimizing the objective w.r.t. $W_{\lambda_{l}}$ can be achieved by performing stochastic gradient descent (SGD) given a fixed $\hat{y}$. We use the above-outlined procedure in an iterative manner (multiple forward, and back-propagation iterations) to align the outcome of the segmentation network with the scribble input provided by the user.

\begin{comment}
\begin{figure*}[t]
%	\setlength\abovecaptionskip{-25pt}
	\includegraphics[width=\linewidth]{images/loss_propagation}
	\caption{The deep learning network provides the initial segmentation. The user then provides scribbles in areas of correction. The loss is then computed over the entire image and this is in turn back-propagated through the network.}
	\label{fig:loss_prop}
\end{figure*}
\end{comment}

\begin{algorithm*}[t]
    \DontPrintSemicolon
	\SetKwInOut{Input}{Input}
	\SetKwInOut{Output}{Output}
	\SetKwInOut{Initialize}{Initialize}
	\SetAlgoLined
	\Input{test instance $x$, input specific scribbles $\mathbb{S}^x$, max epochs $M$, pre-trained weights $W$, $\eta$ learning rate, $\alpha$ regularization factor \\
	  $W_{\lambda} \leftarrow W$	\tcp*{reset to have instance-specific weights}
      }
	\Output{Refined segmentation}
	%\Initialize{$f_{new} = f$}
	\While{\text{$g(y,\mathbb{S}^x) \ > \ 0$ and iteration $< M$ }}
	{
		$y \leftarrow f(x;W_{\lambda}$) \tcp*{perform infererence using weights $W_{\lambda}$}
        $\nabla \leftarrow g(y,\mathbb{S}^x)\frac{\partial{}}{\partial{W_{\lambda_{l}}}} \Psi(x,y,W_{\lambda_{l}}) + \alpha \frac{W_{l} - W_{\lambda_{l}}}{\lvert \lvert W_{l}-W_{\lambda_{l}}\rvert \rvert _{2}}\ $ \tcp*{constraint loss}
       $W_{\lambda_{l}} \leftarrow W_{\lambda_{l}} - \eta \nabla$ \tcp*{update instance-specific weights with SGD}
	}
	\Return{$y$, the refined segmentation}
	\caption{Scribble aware inference for neural networks}
	\label{algo}
\end{algorithm*} 

\cref{fig:proposed_framework} gives a visual description of our framework. It explains the stochastic gradient-based optimization strategy, executed in a manner similar to the standard back-propagation style of gradient descent. However, the difference is that while the back-propagation updates the weights to minimize the training loss, the proposed stochastic gradient approach biases the network output towards the constraints generated by the user provided scribbles at the test time. %Algorithm ~\ref{algo} gives the pseudo-code description of the steps of our overall algorithm.

\mypara{Scribble Region Growing} 
The success of an interactive segmentation system is determined by the amount of burden on a user. This burden can be eased by allowing the user to provide fewer, shorter scribbles. However, providing shorter scribbles can potentially entail a greater number of iterations to obtain the final accurate segmentation. Hence, we propose using region growing to increase the area covered by the scribbles. We grow the region to a new neighborhood pixel, if the intensity of the new pixel differs from the current pixel by less than a threshold $T$.
%and obtain an accurate segmentation with fewer iterations. The extended scribble is denoted as $S_{extended}$, the original scribble as $S_{x}$ and the image as $I$. We consider all the pixels $s$ which are a part of the scribble $S_{x}$ as seed pixels. The criteria to include a pixel $i$ from the image $I$ is given as $\bigcup_{s \in S_{x}} |i - s| < T$ where $s$ denotes the intensity of the image at the location of the scribble, $(x,y)$ include the set of scribble pixels and $T$ is the threshold.

\section{Results and Discussions} \label{results}

\vspace*{-0.25cm}

\myfirstpara{Dataset and Evaluation Methodology}
To validate and demonstrate our method, we have evaluated our approach on the following publicly available datasets containing images captured in different modalities:
\begin{enumerate*}[label=\textbf{(\arabic*)}]
\item \textbf{Microscopy:} 2018 Data Science Bowl (2018 DSB) \cite{datasciencebowl} (nucleus), MonuSeg \cite{monuseg} (nucleus), and ConSeP \cite{hovernet} datasets (epithelial, inflammatory, spindle shaped and miscellaneous cell nuclei) 
\item \textbf{CT:} LiTS \cite{challenlitsge} (liver and tumor cells) and SegThor \cite{segthor} (heart, trachea, aorta, esophagus) challenges 
\item \textbf{MRI:} BraTS' 15 \cite{brats} (necrosis, edema, non-enhancing tumor, enhancing tumor) and CHAOS \cite{CHAOS} (liver, left kidney, right kidney, spleen) datasets. 
\end{enumerate*}
All the experiments were conducted in a Linux environment on a 20 GB GPU (NVIDIA 2018Tx) on a Core-i10 processor, 64 GB RAM, and the scribbles were provided using the WACOM tablet. For microscopy images, the segmented image was taken and scribbles were provided in areas where correction was required using LabelMe \cite{torralba2010labelme}. For CT and MRI scans, the scribbles were provided in the slices of the segmentation scan where correction was desired using 3-D Slicer \cite{pieper20043d}. 
For validating on each of the input modalities, and the corresponding dataset, we have taken a recent state-of-the-art approach for which the DNN model is publicly available and converted it into an interactive segmentation model. We used the same set of hyper-parameters that were used for training the pre-trained model. The details of each model, and source code to test them in our framework are available at \cite{project-page}. To demonstrate the time saved over manual mode, we have segmented the images/scans using LabelMe for microscopy, and 3-D Slicer for CT/MRI, and report it as full human annotation time (F). We took the help of two trained annotators, two general practitioners and a radiologist for the annotation. 

\mypara{Ablation Studies}
We also performed ablation studies to determine : 
\begin{enumerate*}[label=(\alph*)]
	\item Optimum number of iterations, 
	\item Layer number upto which we need to update the weights, 
	\item Type of user input (point,box,scribble) and, 
	\item Effect of scribble length on the user interaction time. 
\end{enumerate*}
Owing to space constraints, the result of the ablation studies are provided on the project page \cite{project-page}. We find scribble as the most efficient way of the user input through our ablation study, and use them in the rest of the paper.

\begin{table}[t]
	\setlength\belowcaptionskip{-50pt}
	\setlength{\tabcolsep}{1.2pt}
	\begin{tabular}{l|ccccccccc|cccccccc}
		\hline
		\multirow{2}{*}{\textbf{Dataset}} & \multicolumn{9}{c|}{\textbf{User Interaction Time}} & \multicolumn{8}{c}{\textbf{Machine Time}} \\
		\cline{2-18}
		& \textbf{F} & \textbf{R} & \textbf{N} & \textbf{\cite{rother2004grabcut}} & \textbf{\cite{fbrs}} & \textbf{\cite{dextr}} & \textbf{\cite{3d-grabcut}} & \textbf{\cite{geos}} & \textbf{\cite{slicseg}}  & \textbf{R} & \textbf{N} & \textbf{\cite{rother2004grabcut}} & \textbf{\cite{fbrs}} & \textbf{\cite{dextr}} & \textbf{\cite{3d-grabcut}} & \textbf{\cite{geos}} & \textbf{\cite{slicseg}} \\
		\hline
		\textbf{\small{2018 DSB}} & 66 & 5 & 7 & 13 & 12 & 12 & - & - & - & 6 & 10 & 11 & 12 & 13 & - & - & - \\
		\textbf{\small{CoNSeP}} & 30  & 6 & 8 & 16 & 18 & 20 & - & - & - & 5 & 7 & 17 & 20 & 23 & - & -  & - \\
		\textbf{\small{LiTS}} & 120 & 7 & 8 & - & - & - & 11 & 12 & 13 & 10 & 12 & - & - & - & 11 & 13 & 11 \\
		\textbf{\small{CHAOS}} & 136 & 13 & 15 & - & - & - & 58 & 66 & 83 & 25 & 30 & - & - & - & 50 & 66 & 83 \\
		\textbf{\small{BraTS' 15}} & 166 & 11 & 13 & - & - & - & 76 & 83 & 100 & 58 & 81 & - & - & - & 100 & 116 & 133 \\    
		\hline
	\end{tabular}
	\caption{User Interaction Time (\textbf{UT}) and Machine Time (\textbf{MT}) in minutes to separate structures (\textbf{F:} Full Human Annotation, \textbf{R:} Our method - Region Growing, \textbf{N:} Our Method - No Region Growing.
		%\textbf{GC:} GrabCut, \textbf{3D-GC:} 3D-GrabCut, \textbf{SS:} SlicSeg). 
		Methods \cite{rother2004grabcut,fbrs,dextr,3d-grabcut,geos,slicseg} were applied till a dice coefficient of 0.95 was reached. }. 
	\label{tab:user_machine_combined}
	\vspace{-1em}
\end{table}

\begin{figure*}[t]
	\setlength{\belowcaptionskip}{-100pt}
	\includegraphics[width = 0.9\textwidth]{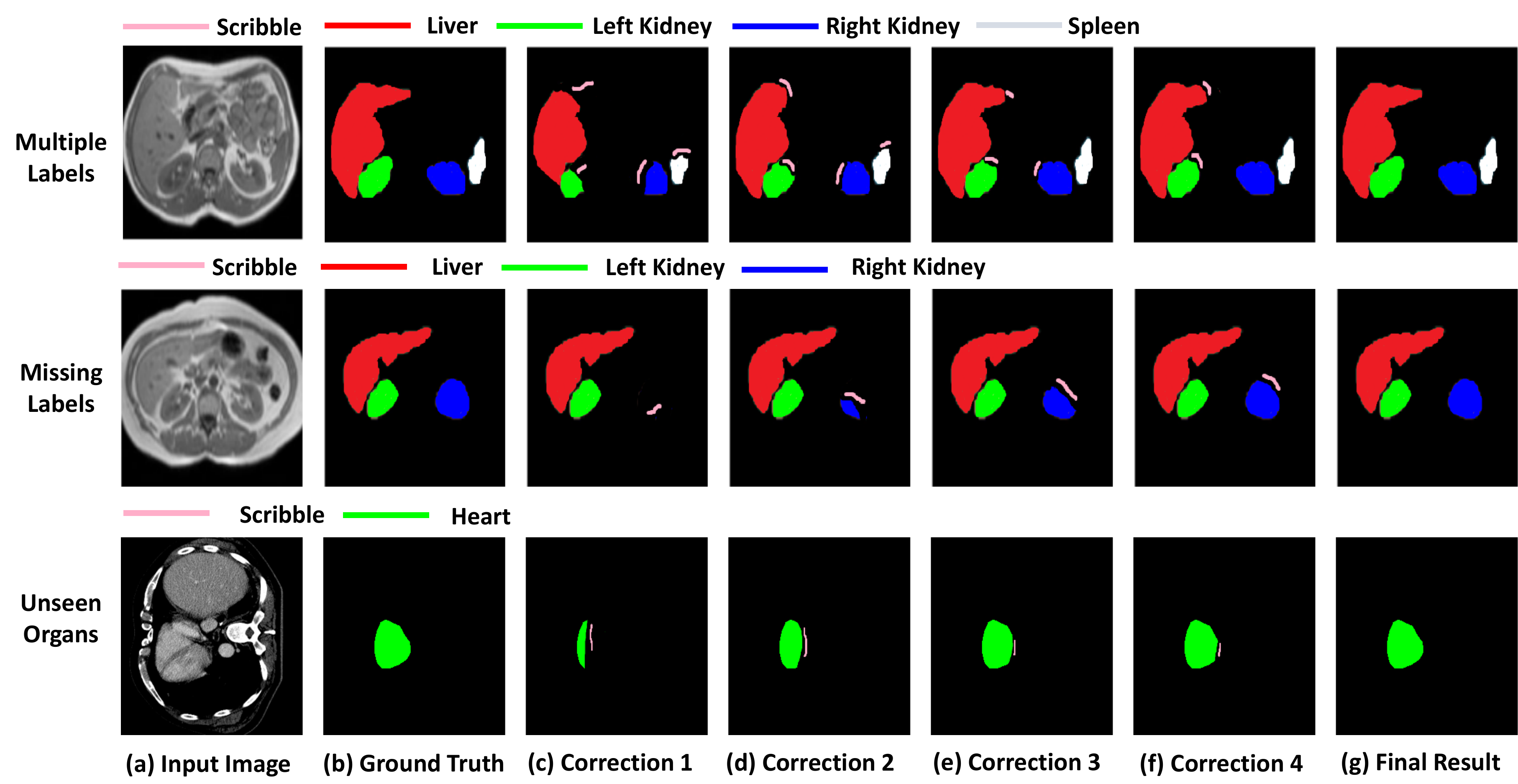}
	\caption{(a) Correcting multiple labels (top row) (b) Inserting missing labels (middle row) (c) Interactive segmentation of organs the model was not trained for (bottom row). Incremental improvement as scribbles are added  shown. No other state-of-the-art approach has these capabilities. More qualitative results are provided here \cite{project-page}.}
	\label{fig:multiple_label}
	
	\vspace*{-0.05cm}
\end{figure*}

\begin{figure*}
\centering
\includegraphics[width=\linewidth]{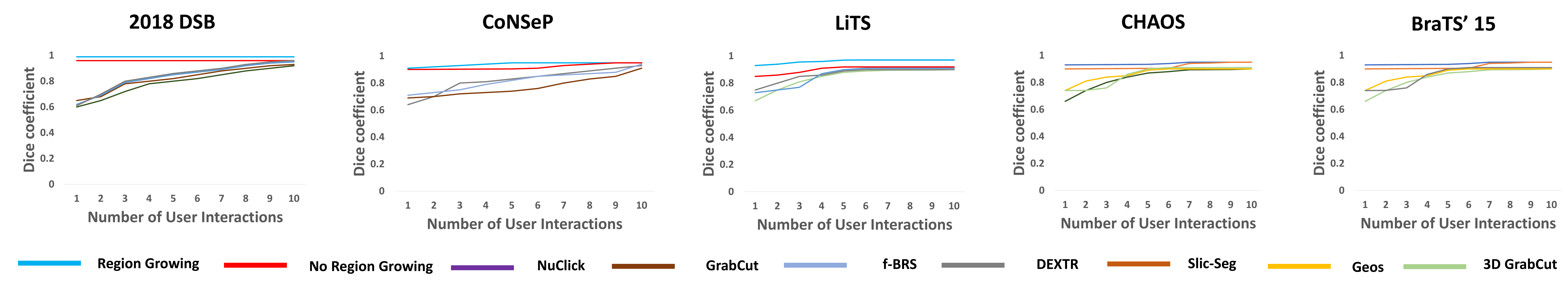}
\vspace{-1em}
\caption{Improvement in segmentation accuracy per user interaction: Our models (region and no-region growing) consistently achieve best accuracy, and in the least number of user interactions.}%
\label{fig:seg_chaos_2018dsb}
\vspace{-1em}
\end{figure*}

\begin{table}
            \footnotesize
              \setlength{\tabcolsep}{3pt}
                \begin{tabular}[b]{lccccc}
\toprule
\textbf{Tissue   Type} & \textbf{1} & \textbf{2} & \textbf{3} & \textbf{4} & \textbf{5} \\
\midrule
\textbf{Nucleus} & 0.54 & 0.62 & 0.76 & 0.81  & 0.86 \\
\textbf{Healthy} & 0.64 & 0.73 & 0.79 & 0.85  & 0.9  \\
\textbf{Necrosis} & 0.61 & 0.65 & 0.72 & 0.81  & 0.85 \\
\textbf{Edema}    & 0.72 & 0.75 & 0.82 & 0.89  & 0.92  \\
\textbf{Enhancing tumor}  & 0.62 & 0.65 & 0.74 & 0.85 & 0.89 \\
\textbf{Non-Enhancing tumor} & 0.71 & 0.75 & 0.83 & 0.87 & 0.92 \\
\textbf{Liver}    & 0.73 & 0.75 & 0.81 & 0.89 & 0.92 \\
\textbf{Tumor}    & 0.67 & 0.72 & 0.83 & 0.87 & 0.89  \\ 
\bottomrule
\end{tabular}
                \hfill
                \begin{tabular}[b]{lcc}

\toprule
\textbf{Method}     & \textbf{\begin{tabular}[c]{@{}c@{}}UT\end{tabular}} & \textbf{\begin{tabular}[c]{@{}c@{}}MT\end{tabular}}  \\
\midrule
\textbf{Ours} & 8 & 7 \\
\textbf{Nuclick} & 13 & 10 \\
\textbf{DEXTR}  & 20 & 11 \\
\textbf{f-BRS}  & 23 & 12 \\
\textbf{GrabCut} & 25 & 13 \\
\bottomrule
\end{tabular}
                
                \caption{\textbf{Left:} Dice Coefficient improvement for tissues with each interaction by medical expert. \textbf{Right:}  User Interaction Time (\textbf{UT}) and Machine Time (\textbf{MT}) for  distribution mismatch scenario (in mins).}
                \label{tab:gt}
%                \vspace{-1.5em}
            \end{table}

\mypara{Image Segmentation with Multiple Classes} 
Our first experiment is to evaluate interactive segmentation in a multi-class setting. We use two trained annotators for the experiment. We have used the validation sets of the 2018 Data Science Bowl (2018 DSB), CoNSeP, LiTS, CHAOS and the BraTS' 15 challenge datasets for the evaluation. We have used the following backbone DNNs to demonstrate our approach: \cite{datasciencebowl, hovernet, pipofan, chaosseg, autofocus}. The details of the networks are provided on the project webpage due to a lack of space. For the microscopy images we compare against Grabcut \cite{rother2004grabcut}, Nuclick \cite{nuclick}, DEXTR \cite{dextr} and f-BRS \cite{fbrs}. For the CT and MRI datasets, we have compared our method against 3-D GrabCut \cite{3d-grabcut}, Geos \cite{geos} and SlicSeg \cite{slicseg}. \cref{tab:user_machine_combined} shows that our technique gives an improvement in user annotation time of 13.3, 12.5, 17.8, 10.2 and 12.4 times compared to full human annotation time and 2.8, 3.0, 1.9, 4.4 and 8.6 times compared to other approaches for nucleus, multiple cells, liver and tumour, multiple organs, and brain segmentation respectively. We also compared the segmentation accuracy per user interaction for every method. Fig. \ref{fig:seg_chaos_2018dsb} shows that our method with region growing outperforms all the methods both in terms of accuracy achieved, and the number of iterations taken to achieve it.

\cref{fig:multiple_label} shows the visual results. The top row shows the segmentation obtained by adding multiple labels in one interaction by our approach. We segment both the tumors and the entire liver by using two scribbles at the same time. One of the important capabilities of our network is to add a label missing from the initial segmentation which is shown in the middle row. Note that our method does not require any pre-training with a specific backbone for interactive segmentation. This allows us to use the backbone networks that were trained for segmenting a particular organ. This ability is especially useful in the data-scarce medical setting when the DNN model for a particular organ is unavailable. This capability is demonstrated in the bottom row of \cref{fig:multiple_label} where a model trained for segmenting liver on LiTS challenge \cite{challenlitsge} is used to segment the heart from SegThor challenge \cite{segthor}. 

\mypara{Distribution Mismatch} 
The current methods cannot handle distribution mismatches forcing pre-training on each specific dataset, requiring significant time, effort, and cost. Our method does not need any pre-training. We demonstrate the advantage on the MonuSeg dataset \cite{monuseg} using the model pre-trained on the 2018 Data Science Bowl \cite{datasciencebowl}. \cref{tab:gt} (Right) shows that our method requires much less user interaction and machine time compared to other methods.

\mypara{Evaluation of our method by medical experts}  
Our approach was tested by medical experts: two general practitioners and a radiologist. We select five most challenging images/scans from the 2018 Data Science Bowl, LiTS, and BraTS' 15 datasets with the least dice score when segmented with the pre-trained segmentation model. The LiTS and the BraTS' 15 datasets were selected owing to their clinical relevance for the diagnosis and volumetric estimation of tumors. 
%Figures \ref{fig:doctor_qualitative} shows the visual results of the improvement in segmentation. We see that our method greatly improves the segmentation of the image by successfully separating the nuclei (top row) and segmenting the entire liver (bottom row) compared to the initial segmentation. We see that incorporating the scribble inputs by the medical experts is useful to improve the segmentation results irrespectively of the modality, mode of capture and the patient. 
\cref{tab:gt} (Left) gives the dice coefficient after each interaction. The improvement in user interaction and machine time are provided in the supplementary material on the project webpage.

\section{Conclusion} \label{conclusion}

\vspace*{-0.25cm}
Modern DNNs for image segmentation require a considerable amount of annotated data for training. %, which is difficult in many medical imaging situations.
%to get for various combinations of different anatomical structures, and image capturing modalities in medical imaging.
Our approach allows using an arbitrary DNN for segmentation and converting it to an interactive segmentation. Our experiments show that we did not require any prior training with the scribbles and yet outperform the state-of-the-art approaches, saving upto 17x (from 120 to 7 mins) in correction time for a medical resource personnel. %which were trained explicitly with the user inputs in terms of user interaction required to reach a near perfect accuracy. %We would like to end with a note that though we have showed our experiments on the supervised approaches, our technique remains valid for the neural network trained in unsupervised or self supervised fashion as well, as long as the accuracy is less than perfect on the test data, and they were originally trained using gradient descent based optimization. We could not demonstrate this aspect of our approach due to unavailability of such techniques in the current literature, which could have been used as backbone within our framework. 

%\bibliographystyle{splncs04}
%\bibliography{sections/references}

\clearpage
\appendix

\begin{center} \textbf{Supplementary Material} \\ \textbf{Efficient and Generic Interactive Segmentation Framework} \end{center}
%
%The supplementary material for our submission to MICCAI 2021 is provided here.

\section{Ablation Study}

\begin{figure*}[!htbp]
	\includegraphics[width = 0.3\linewidth]{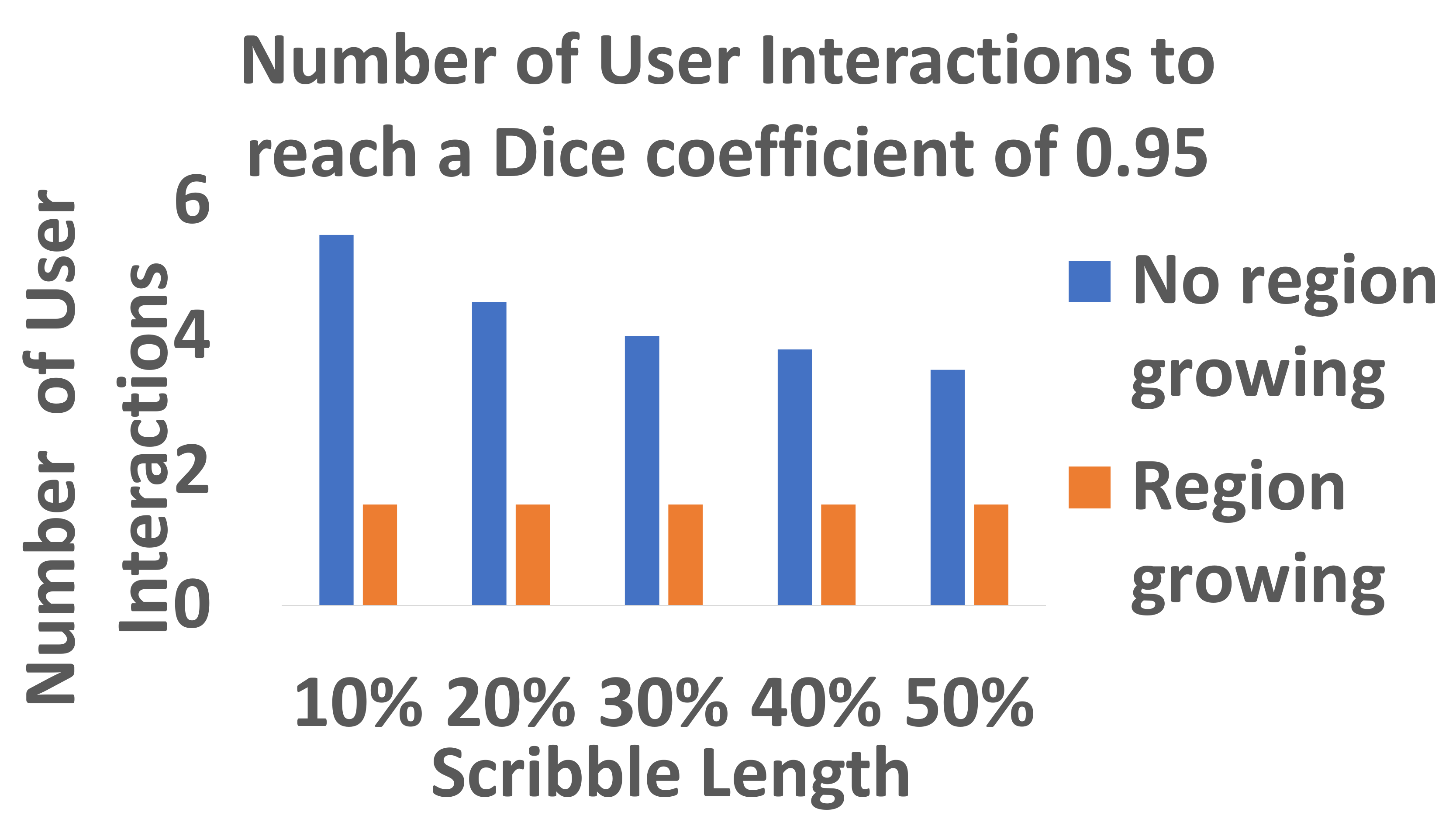}% 
	\quad
	\includegraphics[width = 0.3\linewidth]{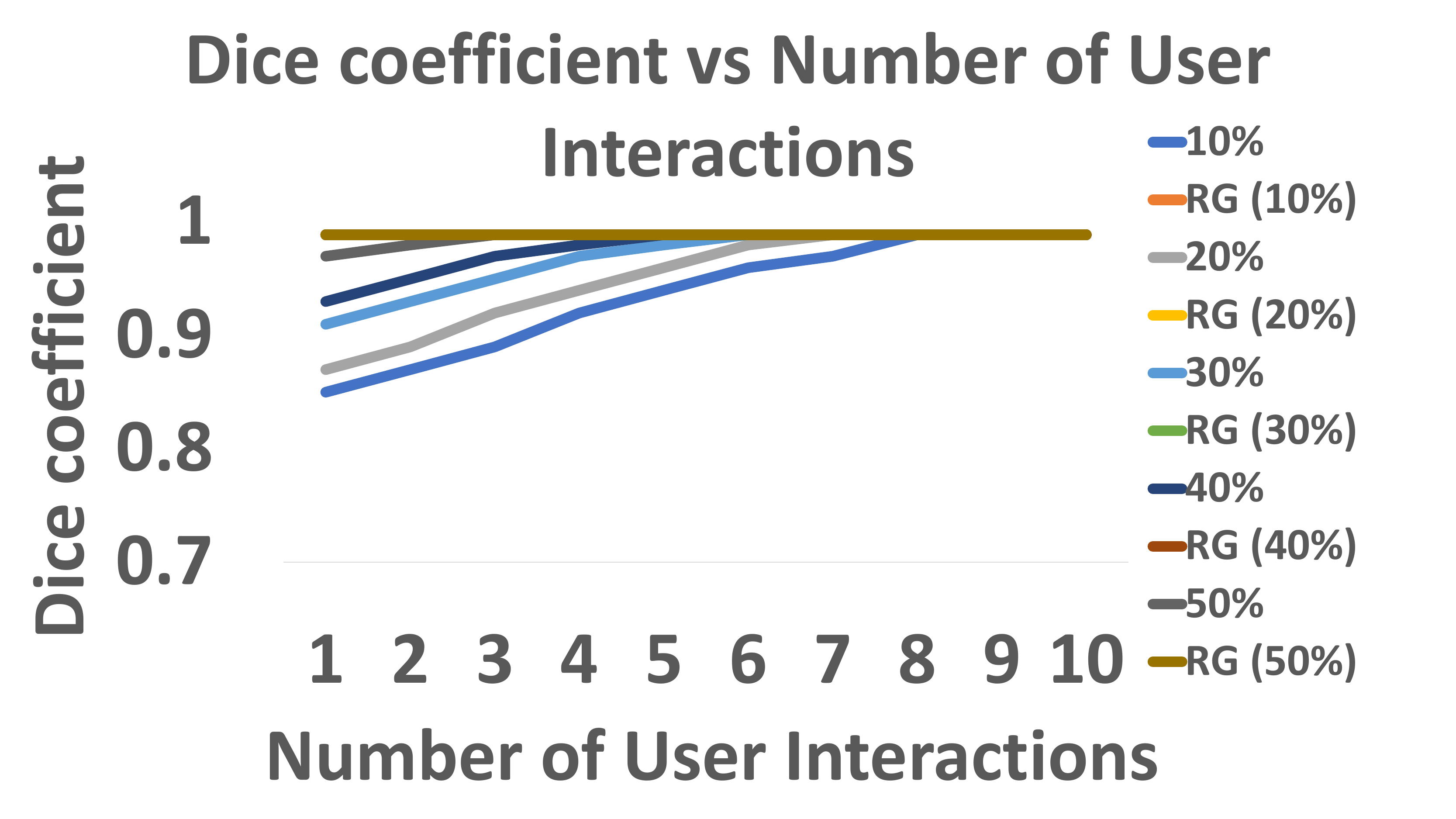}% 
	\quad
	\includegraphics[width = 0.3\linewidth]{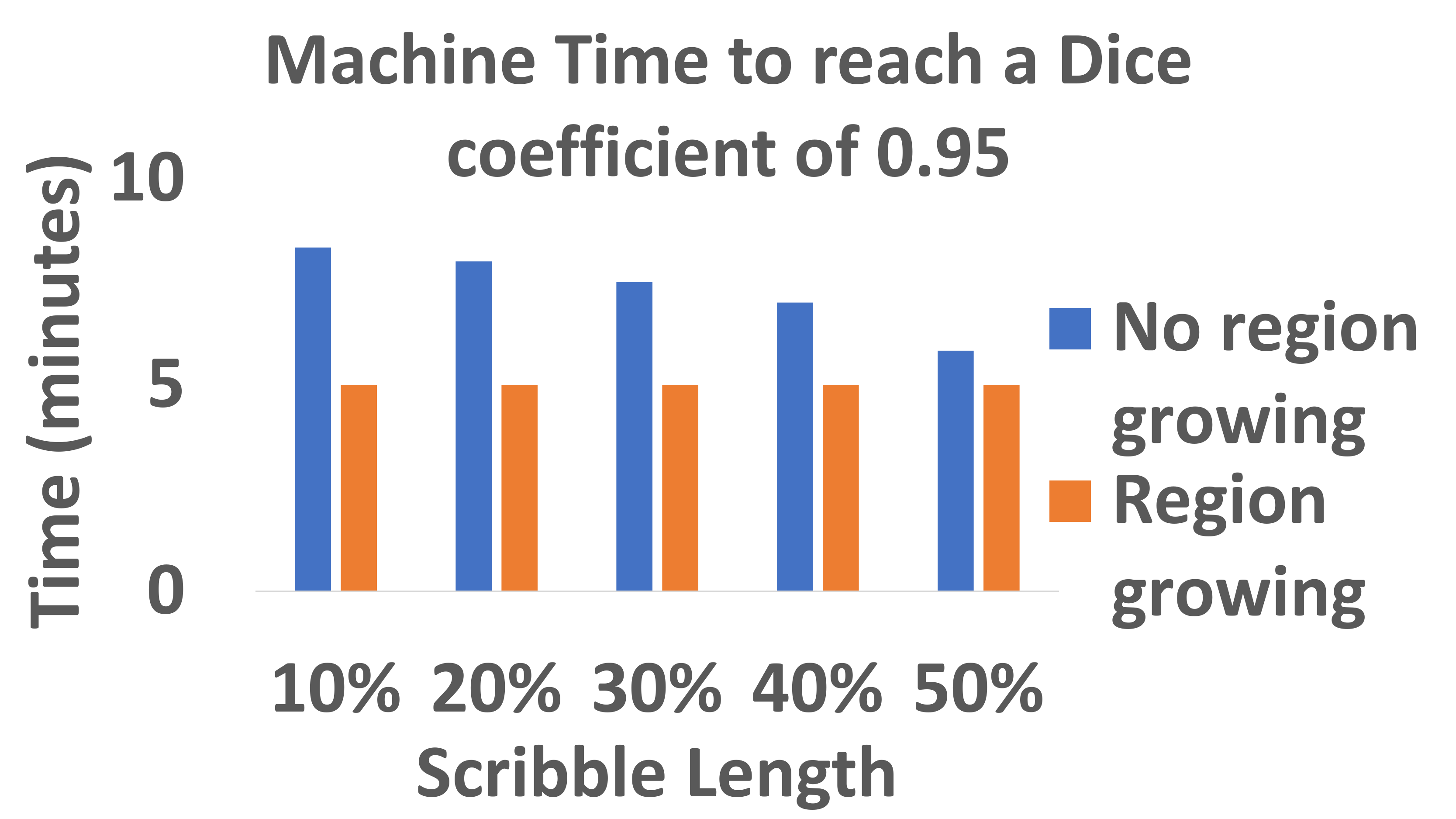}% 
	\vspace{-1em}
	\caption{Effect of scribble length on the increase in the number of user interactions (RG - Region Growing).}%
	\label{fig:region_growing}
\end{figure*}

We conducted studies to determine the effect of various hyper-parameters for our method. 
\begin{enumerate}
\item \textbf{Optimum iterations:} We obtained the optimum number of iterations of back-propagation for obtaining a dice coefficient of 0.95 for each segmentation network. As seen in the Fig. \ref{fig:iteration}, the optimum number of iterations was 80, 100, 130, 140 and 130 for the 2018 Data Science Bowl [5], CoNSeP [8], LiTS Challenge [6], CHAOS dataset [13] and BraTS' 15 [20] segmentation respectively. 

\begin{figure}\CenterFloatBoxes
\begin{floatrow}
\ffigbox[\FBwidth]
{\caption{Optimum Iterations Determination}\label{fig:iteration}}
{\includegraphics[width = 40mm]{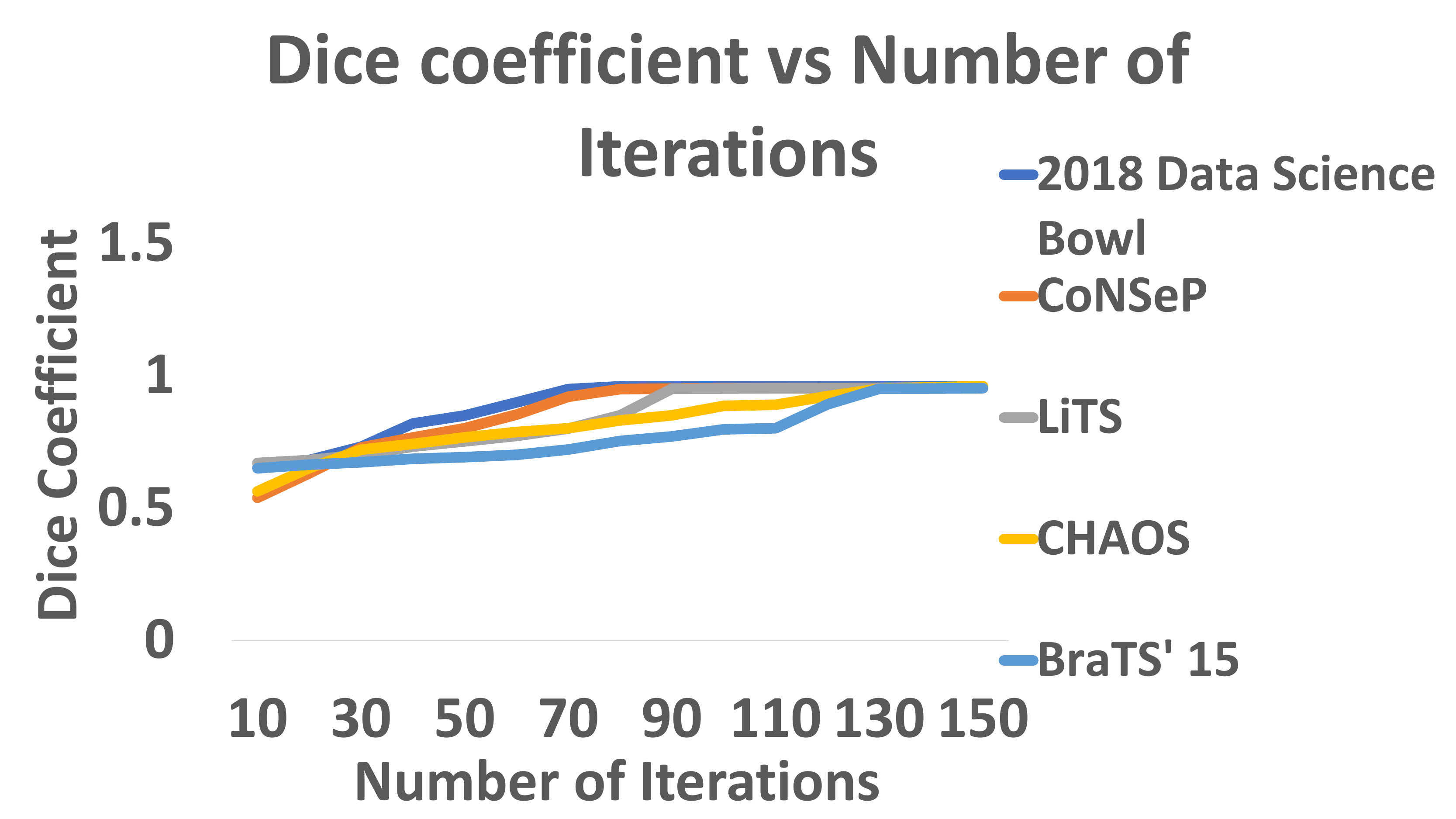}}
\ffigbox[\FBwidth]
{\caption{Optimum Layer Determination}\label{fig:layer}}
{\includegraphics[width = 40mm]{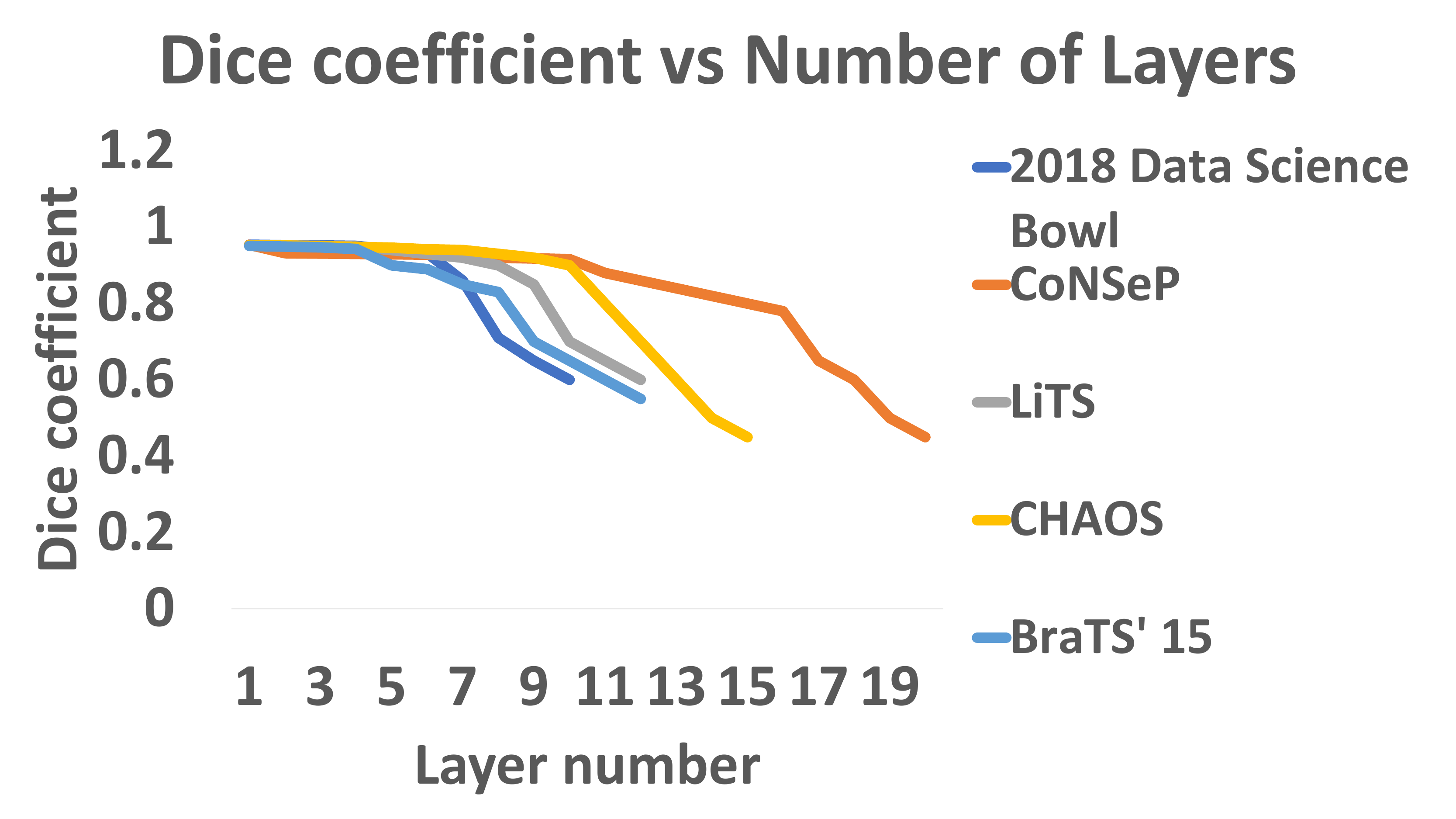}}
\killfloatstyle\ttabbox[0.5\Xhsize]
{\caption{{Number of Mouse Clicks(N), User Time \textbf{(UT)} and Machine Time \textbf{(MT)} for various user inputs (in mins).} }\label{tab:adaptability_user_input}}
{
		\setlength{\tabcolsep}{4pt}
 \begin{tabular}{lccc}
\toprule
\textbf{Method}     & \textbf{\begin{tabular}[c]{@{}c@{}}N\end{tabular}} & \textbf{\begin{tabular}[c]{@{}c@{}}UT\end{tabular}} & \textbf{\begin{tabular}[c]{@{}c@{}}MT\end{tabular}}  \\
\midrule
\textbf{Point} & 100 & 13   & 16  \\
\textbf{Box}  & 34 & 10  & 15     \\
\textbf{Scribble} & 10 & 5 & 10 \\        \bottomrule                            
\end{tabular}

}
\end{floatrow}
\end{figure}

\begin{comment}
\begin{figure*}[!htbp]
    \setlength{\belowcaptionskip}{-100pt}
	\includegraphics[width = 0.45\linewidth]{images/dice_iterations}% 
	\quad
	\includegraphics[width = 0.45\linewidth]{images/dice_layers}
	% 
	\vspace{-1em}
	\caption{\textbf{Ablation Study} The number of iterations (left panel) and the number of layers to back-propagate (right panel) to obtain an optimum dice coefficient have been shown here.}%
	\label{fig:ablation}
\end{figure*}
\end{comment}

\item \textbf{Optimum layer:} Once, the optimum number of iterations are determined, our next step is to determine the optimum layer number upto which back-propagation needs to be performed for each segmentation network. We observe that we obtain the best possible dice coefficient for 4, 6, 4, 3 and 5 layers for the 2018 Data Science Bowl [5], CoNSeP [8], LiTS Challenge [8], CHAOS dataset [13] and BraTS' 15 [20] segmentation respectively as seen in the right panel of Fig. \ref{fig:layer}.
\item \textbf{Optimum user input type:} Our method has the unique and remarkable capability of being able to work with any type of user input such as points, boxes and scribbles. We first performed experiments to determine the most suitable user input modality for segmentation correction. We found that scribbles required the least number of user interactions (30\% lesser mouse clicks), as well as user and machine time (Table \ref{tab:adaptability_user_input}). Hence,  the experiments in the paper were done with scribbles only. 
\item \textbf{Optimum scribble length:} We also evaluated the effect of scribble length while using our method. We observed that without region growing, we needed more user interactions to correct the segmentation as the scribble length reduced. However, with region growing, there was hardly any change in the number of user interactions required as seen in Fig. \ref{fig:region_growing} (obtained for LiTS challenge, similar behavior observed for other datasets, but were not able to provide owing to space restrictions).
\end{enumerate}

\vspace{-3em}

\section{Evaluation of our method by medical expert}

We evaluated our interactive segmentation method with the help of medical experts. We have provided the user interaction time and machine time required for the 2018 Data Science Bowl (2018 DSB), LiTS and BraTS' 15 challenges here. It was possible to obtain a reduction in user annotation time as well as machine time as seen in Table \ref{tab:doctor_user_machine_combined}.

\begin{table}[!htbp]
	\setlength\belowcaptionskip{-50pt}
	\setlength{\tabcolsep}{1.7pt}
	\begin{tabular}{l|ccccccccc|cccccccc}
		\hline
		\multirow{2}{*}{\textbf{Dataset}} & \multicolumn{9}{c|}{\textbf{User Interaction Time}} & \multicolumn{8}{c}{\textbf{Machine Time}} \\
		\cline{2-18}
		& \textbf{F} & \textbf{R} & \textbf{N} & [25] & [28] & [19] & [21] & [32] & [33]  & \textbf{R} & \textbf{N} & [25] & [28] & [19] & [21] & [32] & [33] \\
		\hline
		\textbf{\small{2018 DSB}} & 55 & 4 & 6 & 15 & 14 & 14 & - & - & - & 7 & 12 & 19 & 18 & 15 & - & - & - \\
		\textbf{\small{LiTS}} & 100 & 6 & 7 & - & - & - & 14 & 15 & 16 & 9 & 10 & - & - & - & 12 & 14 & 15 \\
		\textbf{\small{BraTS' 15}} & 150 & 9 & 12 & - & - & -  & 65 & 75 & 90 & 50 & 80 & - & - & - & 120 & 126 & 145 \\    
		\hline
	\end{tabular}
	\caption{User Interaction Time (\textbf{UT}) and Machine Time (\textbf{MT}) in minutes for separating structures by a medical expert (\textbf{F:} Full Human Annotation, \textbf{R:} Our method - Region Growing, \textbf{N:} Our Method - No Region Growing. All the semi-automated methods [19, 21, 25, 28, 32, 33] were applied till a dice coefficient of 0.95 was reached.}. 
	\label{tab:doctor_user_machine_combined}
	\vspace{-1em}
\end{table}
%\vspace{-1em}
%\section{Comparison of our work with Wang et al. \cite{bifseg}}

%Our method is significantly different from Wang et al.: (a) DNN - Wang et al. use their own custom neural networks for interactive segmentation. However, our method can use pre-existing segmentation networks. This allows our method to use new segmentation architectures which may be proposed in the future as well. (b) Optimization - Wang et al. use CRF based regularization for label correction. We propose a novel restricted lagrangian based formulation. This enables us to do a sample specific fine-tuning of the network, and allows our method to do multiple label corrections in a single iteration. This is a novel capability. (c) User Inputs - Wang et al. use scribbles and bounding boxes as user inputs. We can do label correction irrespective of the type of user input provided which is unique.

\end{document}